\documentclass{article}

\usepackage{PRIMEarxiv}

\usepackage[utf8]{inputenc} 
\usepackage[T1]{fontenc}    
\usepackage{hyperref}       
\usepackage{url}            
\usepackage{booktabs}       
\usepackage{amsfonts}       
\usepackage{nicefrac}       
\usepackage{microtype}      
\usepackage{lipsum}
\usepackage{fancyhdr}       
\usepackage{graphicx}       
\graphicspath{{media/}}     

\usepackage{cite}
\usepackage{amsmath,amssymb,amsfonts}
\usepackage{graphicx}
\usepackage{textcomp}
\usepackage{xcolor}

\usepackage{comment}
\usepackage{pgf}
\usepackage{layouts}
\usepackage{caption}
\usepackage{subcaption}
\usepackage{soul}
\usepackage{hyperref}
\hypersetup{draft}
\usepackage{color, colortbl}
\definecolor{Gray}{gray}{0.9}

\title{\textbf{LEAD1.0}: A \textbf{L}arge-scale annotated dataset for \textbf{E}nergy \textbf{A}nomaly \textbf{D}etection in Commercial Buildings
}

\author{
  Manoj Gulati \\
  Singapore Management University\\
  Singapore\\
  \texttt{manojg@smu.edu.sg}
  \And
  Pandarasamy Arjunan\\
  Berkeley Education Alliance for\\ 
  Research in Singapore Limited\\
  Singapore\\
  \texttt{samy@bears-berkeley.sg}
}

\begin{document}
\maketitle

\begin{abstract}
Modern buildings are densely equipped with smart energy meters, which periodically generate a massive amount of time-series data yielding few million data points every day. 
This data can be leveraged to discover the underlying loads, infer their energy consumption patterns, inter-dependencies on environmental factors, and the building's operational properties. 
Furthermore, it allows us to simultaneously identify anomalies present in the electricity consumption profiles, which is a big step towards saving energy and achieving global sustainability. 
However, to date, the lack of large-scale annotated energy consumption datasets hinders the ongoing research in anomaly detection. 
We contribute to this effort by releasing a well-annotated version of a publicly available \emph{ASHRAE Great Energy Predictor III} data set containing 1,413 smart electricity meter time series spanning over one year.
In addition, we benchmark the performance of eight state-of-the-art anomaly detection methods on our dataset and compare their performance.
\end{abstract}


\section{Introduction}
\label{sec:introduction}
Buildings are one of the largest energy consumers, accounting for approximately 40\% of the total energy usage in the world~\cite{shaikh2014review}. It is estimated that 20\% of the total energy consumed gets wasted within buildings~\cite{roth2005energy}. Further, the energy demand of the buildings is increasing continuously and will rise by 28\% by 2040~\cite{conti2016international}. Hence there is a pressing need to reduce energy wastage to lower the energy footprint and cost of buildings' utilities. The most common causes of energy wastage are equipment failure, aging, misconfiguration, and human-related errors. This wasteful use of energy can be identified \& prevented using a data-driven analytical technique known as anomaly detection~\cite{himeur2021artificial}.

More recently, the fast-paced proliferation of smart meters has led to a boost in research focused on leveraging smart meter data for anomaly detection in buildings~\cite{arjunan2015multi}. This surge in anomaly detection research is due to (i) the realization of the significance of the contribution of buildings towards gross energy consumption and (ii) the benefits it can garner in terms of long-term energy sustainability in buildings. Also, it is the most efficient mode of sensing and detection due to ease of installation, monitoring, and scalability. Anomaly detection in building energy consumption is the process of identifying unusual energy usage events that lead to energy waste. Such abnormal use differs significantly from the normal energy usage patterns. The proliferation of advanced metering infrastructure (AMI), along with improved computationally intelligent methods, makes it possible to develop automated anomaly detection techniques~\cite{himeur2021artificial}. These techniques track the building's energy consumption patterns from the aggregate smart meter data, identify anomalous events of energy consumption and report them to building managers for further action.

A vast literature on anomaly detection techniques exists for time series in various domains~\cite{choi2021deep}. In~\cite{himeur2021artificial}, the authors describe some of the practical challenges in detecting anomalies in energy consumption data and reviewed different approaches namely (a) statistics-based rule sets, (b) unsupervised, and (c) supervised. While anomaly detection in building energy consumption is an active area of research, several challenges exist and hampers its acceptance in the real world as a method to screen and optimize energy utilization~\cite{himeur2021artificial}. This include (a) difficulty in collection \& assignment of anomaly labels, (b) lack of annotated public dataset for anomaly detection research, furthermore, (c) due to a lack of annotated dataset, people have widely used unsupervised methods, which lead to a higher rate of false positives. 
Consequently, building operators have to go through all identified anomalies and filter out the genuine ones to take further actions, which is a tedious and time-consuming task. 
Due to this limitation, existing studies have evaluated the models on a few buildings, making it difficult to estimate the true energy-saving potential of the models.

To mitigate these limitations, we annotate and release 
\textbf{LEAD1.0 -- a Large-scale Energy Anomaly Detection}\footnote{https://github.com/samy101/lead-dataset} dataset consisting of 1,413 smart electricity meter data spanning over an year.  
To the best of our knowledge, this dataset is so far the largest for energy anomaly detection in the public domain. 
We also benchmark the performance of several state-of-the-art approaches and release the code-base of anomaly annotation tool in open-source for community use.

The rest of the paper is organized as follows.
In \textbf{Section}~\ref{sec:dataset}, we describe the dataset details and anomaly annotation procedure. In \textbf{Section}~\ref{sec:results}, we present our benchmarking study and results. Finally, in \textbf{Section}~\ref{sec:discussion}, we discuss future directions and conclude this paper. 

\begin{figure*}[t!]
	\centering
	\includegraphics[width=\textwidth]{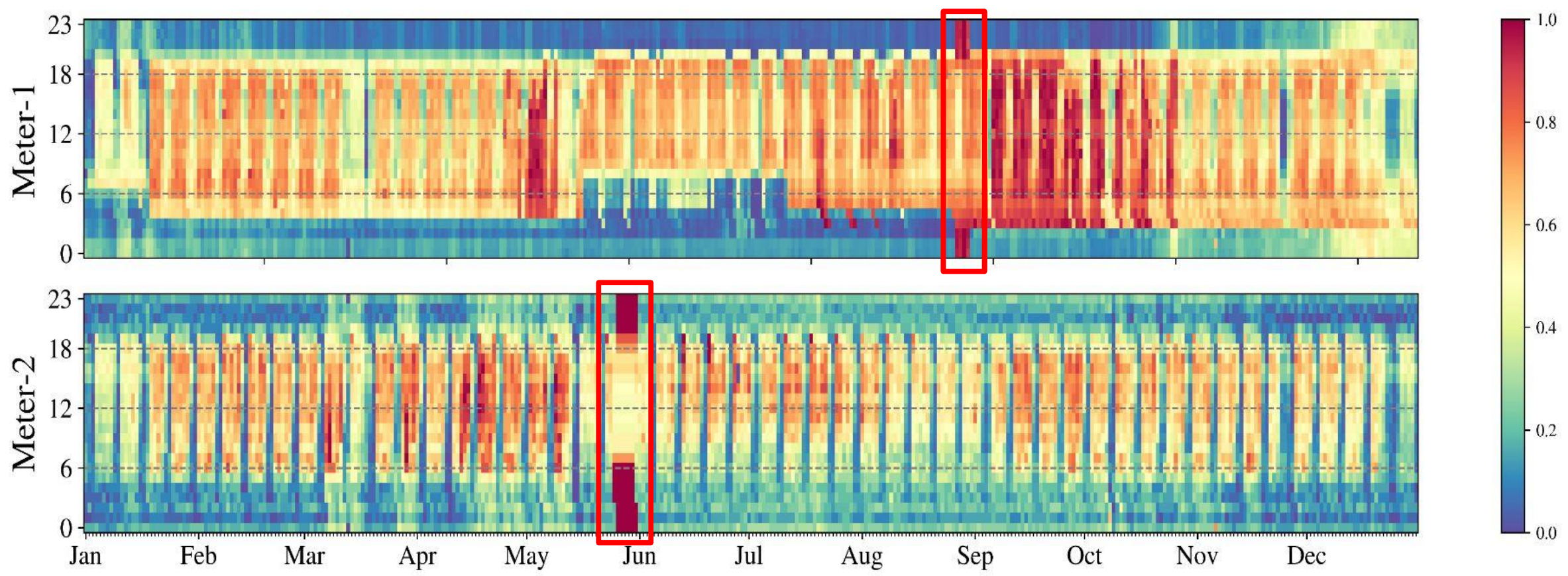}
	\caption{The hourly electricity usage patterns (normalized) of two buildings with examples for point and sequential anomalies. The X-axis denotes the day of the year, and Y-axis is the hour of the day. The colors represent the electricity usage at that hour, with blue being lower and red being higher usage.}
	\label{fig:heatmap}
\end{figure*}

\section{Data set preparation}
\label{sec:dataset}
We leveraged the dataset used in the Great Energy Predictor III competition\footnote{https://www.kaggle.com/c/ashrae-energy-prediction/} conducted in 2019 on the Kaggle platform~\cite{Miller2020-qw}. This dataset includes one year of hourly meter readings from 1,636 non-residential buildings collected from 16 different sites worldwide~\cite{Miller2020-yc}. Also, it contains building meta-data like ${square\_feet}$, ${year\_built}$, and ${floor\_count}$ to describe the structure of the building (specified by the ${building\_id}$). Furthermore, it is accompanied by various weather parameters to help model the buildings' energy usage better. 

This dataset had measurements taken from four different energy meter types (electricity, chilled water, steam \& hot-water). 
For the task of anomaly detection, we exercised hourly meter readings data from 1,413 electricity meters covering 16 different building types, such as office, monitored for one year. %
Please note that in the original ASHRAE dataset, there were 1,636 buildings (not meters). Each building had different energy meter types such as electricity, hot-water, etc. In this paper, we have focused only on the electricity meters (1,413) and in future we plan to annotate other meter types as well.
The top five winners of this competition have annotated some outliers that they excluded for model development; however, these annotations were not comprehensive and were missing class labels for different types of anomalies. 

In our expedition, we annotated this dataset with (a) point anomalies and (b) sequential or collective anomalies:

\begin{enumerate}
\item \textbf{Point anomaly}: A point anomaly refers to an energy consumption instance that appears unusual when compared to the overall/whole time series (global) or compared to its neighboring points (local). It occurs once at any time and does not repeat.
\item \textbf{Sequential or collective anomaly}: A sequential anomaly refers to a consecutive set of energy consumption events whose joint behavior is unusual. It may occur once or repeatedly at regular intervals. Sequential anomalies can also be local or global. 
\end{enumerate}

\begin{figure}[t]
	\centering
	\includegraphics[scale=0.5]{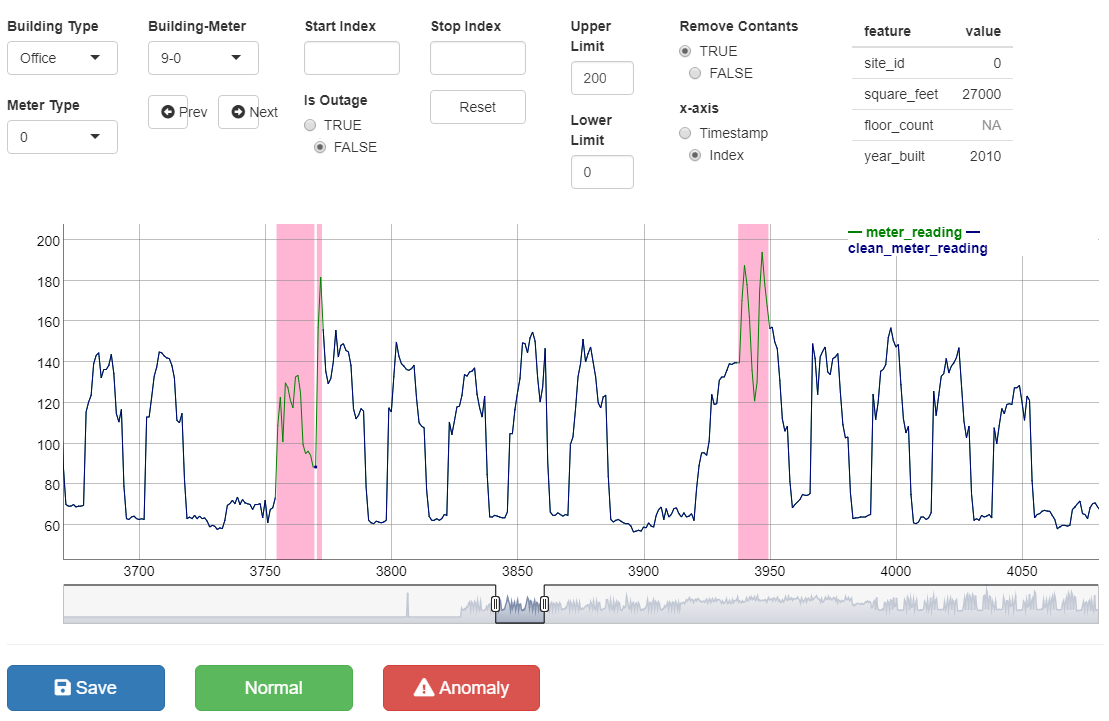}
	\caption{The user interface of our web-based anomaly annotation tool for energy time series.}
	\label{fig:anomaly_annotator}
\end{figure}

An example of these anomalies from our dataset are shown in Figure~\ref{fig:heatmap}. We have developed a web-based tool (shown in Figure~\ref{fig:anomaly_annotator}), to annotate every point in the electricity meter time series. This process involved manually examining approximately 12 million data points in total, with each inspection window having 8,784 data points on an average for each electricity meter.
The original ASHRAE dataset contains hourly readings for the entire year of 2016. Since 2016 is a leap year, there were 8,784 (366x24) samples for each meter. 
Each manual inspection (over 24 hour window span) took somewhere from three to five minutes, collectively accounting for approximately 100 man-hours invested for this strenuous annotation exercise.
Please note, that we have carefully annotated with an hourly data span, and not full days. The current benchmarking study (discussed in Section-3) is focused on detecting whether a day has any anomaly or not, inline with some previous studies found in the literature. However, it is quite possible to pinpoint the anomalies in our dataset using a different experimental setup with the same PyOD models.

%
We followed a fixed protocol for annotating each time window (as discussed below). For each meter's time-series plot, we zoom in and look for a weekly or daily pattern in the time series of meter readings.
(i) If the plot follows a weekly or daily pattern, then we look for the disturbance in this pattern. And that disturbance is marked as an anomaly.
(ii) If the plot does not have any suitable pattern, then we look for the days having higher or lower energy consumption than the usual. If we find that there is a large difference between that day’s energy consumption and its nearby days, then we mark that as an anomaly. It is hard to follow the same set of rules for annotating anomalies in all the buildings because each building has a different definition of anomaly but above mentioned steps are the most common steps that we applied while doing annotation. Also, the tool that we have created provides us a better and efficient way to annotate. We have released this web-tool along with this dataset for the public use. 
Please note, these anomalies were defined based on literature and after going through each raw time series and also enforced a verification process as part of our anomaly annotation protocol. 
%
%
Precisely, we had 199,640 anomalous instances present across 12,060,910 data points logged by 1,413 energy meters (excluding the missing data points which would make the total count of 12,411,792 data points i.e. 1,413 meters x 8784 [366days x 24hrs]).

\begin{table*}[t!]
	\centering
	\caption{Comparison of F1-score, Precision and Recall of all anomaly detection models.}
	\begin{tabular}{lrrr}
	\toprule
		\textbf{Model} & \textbf{F1-score} & \textbf{Precision} & \textbf{Recall} \\\midrule
		Cluster-based Local Outlier Factor (CBLOF) & 0.425 & 0.900 & 0.277 \\ 
		Feature Bagging & 0.424 & 0.899 & 0.279 \\ 
		\cellcolor{Gray} K-Nearest Neighbors (KNN) & \cellcolor{Gray} 0.431 & \cellcolor{Gray} 0.902 & \cellcolor{Gray} 0.284 \\
		Histogram-base Outlier Detection (HBOS) & 0.397 & 0.896 & 0.258 \\ 
		Isolation Forest & 0.413 & 0.895 & 0.270 \\ 
		One-class SVM (OCSVM) & 0.421 & 0.899 & 0.276 \\ 
		Local Outlier Factor (LOF) & \cellcolor{Gray} 0.426 & 0.900 & \cellcolor{Gray} 0.281 \\ 
		Minimum Covariance Determinant (MCD) & 0.422 & \cellcolor{Gray} 0.901 & 0.276 \\ 
		\bottomrule
	\end{tabular}
	\label{tab:tab1}
\end{table*}

\section{Benchmarking}
\label{sec:results}
\textbf{Data pre-processing:} We first create new categorical features based on the timestep - hour, day, weekday, and month. Then, we normalize all the available features using the z-score normalization technique. We don't directly use the meter-reading data but calculate its natural logarithm i.e. log(p+1).  
We have used log(p+1) for data-normalization, as this is a standard way to scale the target before fitting the model and use it for prediction purposes.
Post data-normalization, we group the data available for each building and apply a sliding window across the available timesteps. In this paper, we have experimented with 24-hour windows with zero overlaps, so that models try to learn daily energy patterns for each building. 

\textbf{Benchmarking Training Models}: We have considered a variety of statistics-based methods as well as machine learning-based methods for benchmarking anomaly detection on our dataset. Conventionally, clustering-based algorithms are already being employed in various applications where anomaly detection has been studied. In our work, we evaluate some of these techniques, which include (a) Cluster-based Local Outlier Factor (CBLOF), (b) Feature Bagging, (c) Histogram-base Outlier Detection (HBOS), (d) Isolation Forest, (e) K-Nearest Neighbors (KNN), (f) Local Outlier Factor (LOF), (g) Minimum Covariance Determinant (MCD), and (h) One-class SVM (OCSVM). All these models were implemented in Python with the help of the PyOD library\footnote{https://github.com/yzhao062/pyod} and executed on Google Colab platform. 

\textbf{Evaluation Metrics:} In this work, we have used Precision, Recall, and F1 score to have a fair understanding of the distribution of false positives and false negatives while determining anomalous profiles.  
The formulas are as follows -
\begin{equation} \label{eq}
	Precision = \frac{True Positives}{ True Positives + False Positives} 
\end{equation}
\begin{equation} \label{eq}
	Recall = \frac{True Positives}{ True Positives + False Negatives} 
\end{equation}
\begin{equation} \label{eq}
	F_1 score = \frac{2 * Precision * Recall}{ Precision + Recall}
\end{equation}

\textbf{Experimental setup:} Firstly, we separate the anomalous and the non-anomalous data. The non-anomalous samples were split into train(80\%), validation(10\%) and test(10\%) sets, and the anomalous samples were  split into train(10\%), validation(20\%) and test(70\%) sets. Both the non-anomalous and anomalous sets were combined to make the final training, validation, and test tests. 
The model is trained on the training sets and then the validation set. The threshold calculated based on the validation sets is then used on the test sets to measure the model performance.

\textbf{Model comparison}:
Please note that we are not interested to specifically pinpoint the time point of the anomalous activity. 
However, in this work, we focus on identifying whether the entire 24-hour sequence is anomalous or not. 
We compute F1-Score, Precision and Recall (as mentioned previously) on our annotated dataset. 
These results are presented in Table~\ref{tab:tab1}. 
K-Nearest Neighbors and Minimum Covariance Determinant yield the highest precision i.e. 0.902 and 0.901 respectively.
In terms of recall, KNN stands out by yielding 0.284 recall followed by Local Outlier Factor whose recall is around 0.281.
In terms of F1-score, obviously KNN outperforms all the models by yielding an F1-score of 0.431, followed by Local Outlier Factor whose score is around 0.426.
We acknowledge that we have only tried to conduct a limited benchmarking exercise here, and the list of models that we have experimented with is no where an exhaustive list and definitely there are plenty of other models which can tried out on this data.
%

\section{Discussion and Conclusions}
\label{sec:discussion}
Data-driven energy sustainability in buildings is considered to be a major leap in reducing carbon footprint and assert the pace of the ongoing climate change. 
Anomaly detection is one such technique, which caters to this by identifying avoidable electricity usage and reporting it to the stakeholders.
In this work, we annotate a large public dataset having electricity data from 1,413 non-residential buildings to facilitate the design and evaluation of machine learning techniques for anomaly detection.
%
%
Furthermore, this large dataset collected across different types of buildings (spread across the world) will amplify the development of more generalized machine learning models.
Also, to benchmark \& gauge the correctness of our labels, we evaluated anomaly detection accuracy in terms of F1-Score, Precision, and Recall (shown in Table~\ref{tab:tab1}) using off-the-shelf models.
Apart from this, we released an easy-to-use web tool, which can be used for annotating any time series data. 
Although, this annotated dataset (one year long), along with initial accuracy reports with off-the-shelf models, is a good starting for the development of anomaly detection models. However, this annotation exercise can be further extended to other public datasets.
Also, in the future, we plan to exercise more comprehensive models using state-of-the-art deep learning techniques to improve the efficacy of anomaly detection further.
We aim to release these findings in our future work, which is planned along these lines.

\section*{Acknowledgments}
The authors acknowledge the planning and technical committee members of the \emph{ASHRAE - Great Energy Predictor III} for hosting and releasing this dataset. We also acknowledge Bhavuk Singhal and Mudit Dhawan for their contribution towards generating annotations and assistance in establishing the benchmarks.

\bibliographystyle{unsrt}  
\bibliography{references}  

\end{document}